# Temporarily Unavailable:

# Memory Inhibition in Cognitive and Computer Science


Tobias Tempel[1,*], Claudia Niederée[2], Christian Jilek[3,4], Andrea Ceroni[2], Heiko Maus[3], Yannick Runge[5], and Christian Frings[5]

[1] Ludwigsburg University of Education, Reuteallee 46, 71634 Ludwigsburg, Germany

[2] L3S Research Center, University of Hannover, Appelstr. 9a, 30167 Hannover, Germany

[3] German Research Center for Artificial Intelligence (DFKI), Trippstadter Str. 122, 67663 Kaiserslautern, Germany

[4] TU Kaiserslautern, Erwin-Schrödinger-Straße 52, 67663 Kaiserslautern, Germany

[5] Fachbereich I – Psychologie, University of Trier, 54286 Trier, Germany

[*] Corresponding author: tobias.tempel@ph-ludwigsburg.de


– Draft –






**Abstract**

Inhibition is one of the core concepts in Cognitive Psychology. The idea of inhibitory mechanisms actively weakening representations in the human mind has inspired a great number of studies in various research domains. In contrast, Computer Science only recently has begun to consider inhibition as a second basic processing quality beside activation. Here, we review psychological research on inhibition in memory and link the gained insights with the current efforts in Computer Science of incorporating inhibitory principles for optimizing information retrieval in Personal Information Management. Four common aspects guide this review in both domains: 1. The purpose of inhibition to increase processing efficiency. 2. Its relation to activation. 3. Its links to contexts. 4. Its temporariness. In summary, the concept of inhibition has been used by Computer Science for enhancing software in various ways already. Yet, we also identify areas for promising future developments of inhibitory mechanisms, particularly context inhibition.

## 1. Introduction

The term 'inhibition' has been widely used in Psychology to describe a plethora of phenomena. Inhibition can take place at the level of neurotransmitters in the synaptic cleft, neurons can inhibit each other's fire rate, it can be shown at a physiological level – for instance by measuring the EEG, and finally it can be investigated on a purely behavioral level. Behavioral inhibition typically means something like 'making a content/action less accessible or suppressing it altogether' in order to enhance processing of relevant information. In cognition, thus, the concept of inhibition implies cognitive mechanisms that actively lower currently irrelevant or interfering information. Psychological theories that posit the existence of inhibitory mechanisms in our mind have elicited much research across diverse fields of Cognitive Psychology like perception, attention, action control, and memory but have also been transferred to other research fields like Developmental Psychology as, for instance, understanding the aging brain or the developing brain is closely linked to understanding how the brain handles irrelevant or interfering information – that is how or whether the brain can inhibit such information.

The two areas in Cognitive Psychology in which inhibition is traditionally investigated to the largest extent are the research fields of attention and memory. In attention research, typically the interference due to distracting stimuli or actions is analyzed in experimental paradigms that try to tap a specific form of cognitive inhibition. For example, in the *Negative Priming* task (for a review, Frings, Schneider, & Fox, 2015) it is typically analyzed how an irrelevant distractor stimulus is inhibited. In the cuing task that elicits the *inhibition of return* effect (Posner, Choate, Rafal, & Vaughn, 1985) it is typically analyzed how an irrelevant location is inhibited. In task switching (Kiesel et al., 2010) lowering competition by a just previously performed task while currently executing a novel task is achieved by inhibiting that previous task. Finally, in the go-no-





go-task prepotent motor responses have to be inhibited (e.g., Logan & Cowan, 1984).

While the results from inhibition research in attention and action control has clear implications for applied research (e.g., implications for traffic behavior), the purpose of this article is to link the concept of inhibition in Cognitive Psychology to the concept of inhibition in Computer Science. By definition this type of inhibition is thus concerned with inhibiting representations of information – in other words this article focuses on inhibition in memory. Our approach also focuses on output variables as to understand inhibition, that is, we look at human and artificial representational systems and how input variables or internal processes can be modulated by making information already represented in the system less accessible thereby enhancing output performance. For this conjoint approach we thus look at inhibition from a behavioral perspective.

Yet, we should right from the start acknowledge that the concept of inhibition – partially because it is used in so many different areas of research – was often subject to a heated debate (see e.g., volumes by Dagenbach & Carr, 1994; Brainerd & Dempster, 1995; Gorfein & MacLeod, 2007). Researchers argue for decades whether cognition can be explained and described without using the concept of inhibition – and in many research areas this debate is still not consensually solved. For the purpose of this article, however, we think that the debate about the concept of inhibition is secondary as the behavioral outcomes (that presumably were elicited by inhibitory functions) are the central aspect of our approach. Therefore, we do not provide an extensive overview over the evidence that has been reported in favor or against inhibitory functions but focus on carving out what is essential to the concept of inhibition in memory.

## 2. Inhibition in memory: A psychological perspective

In memory, inhibition results in forgetting. Inhibitory processes weaken stored





information. Crucially, this occurs in an active fashion: Inhibition in memory means that forgetting is not only a mere by-product of storing new or strengthening existing representations, causing impaired access to non-strengthened representations, but can reflect the effects of processes that purposefully decrease a representation's current activation level. Moreover, these effects go beyond the classic distinction between accessibility and availability. Tulving and Pearlstone (1966) introduced these concepts to memory theories by demonstrating that memory performance crucially depends on suitable retrieval cues. The present cues determine the accessibility of a trace whereas availability is independent from cues. Thus, a presently inaccessible memory trace may be available and, thus, turn accessible when the right cue is provided.

Inhibition affects availability because it weakens representations independently from associated cues. Cue independence is an essential feature of forgetting effects that have been attributed to inhibition. When memory performance remains impaired (relative to a defined baseline measure) independently of which cues are presented, such a pattern is interpreted as inhibitory mechanisms impacting an individual memory trace's activation level. Inhibition makes thus traces unavailable. Yet, inhibition effects are typically not assumed to be permanent. A release from inhibition after a certain period of time, in fact, is a standard finding and corresponds to the adaptive purpose of inhibition of momentarily improving processing efficiency by weakening currently irrelevant and interfering information (that may become relevant and important in a novel context again). This temporariness goes beyond the classic concept of availability that never has been considered to be able to recover over time. Once a memory became unavailable, it was considered to be lost, unless re-encoded. The concept of inhibition, however, specifies that unavailability can be reversible as inaccessibility by definition





is. Unavailability does not necessarily mean that something was deleted. Inhibition makes memories temporarily unavailable.

Several aspects characterize the concept of inhibition. We focus on four of such aspects, each of which defines inhibition in a cognitive as well as in a computer system: its purpose for increasing processing efficiency, its relation to activation, its links to mental or tasks contexts, and its temporariness. We will first summarize what is known with respect to these four aspects of inhibition in memory before we discuss and transfer these concepts to inhibition in Computer Science.

## 2.1 Processing efficiency

In general, inhibition is considered a basic processing quality beside excitation, responsible for more refined as well as more efficient processing than could result from mere spreading activation among cognitive representations. Inhibition facilitates focusing on relevant information. Any information that is irrelevant with regard to current internal or external demands but nevertheless becomes either cognitively represented through perception or activated within memory can absorb resources and, thus, impair processing aimed at achieving current action goals. Typically, inhibition is closely linked to interference. When currently irrelevant information distracts from accomplishing a task, inhibiting that information resolves interference.

Although memory research suggests that inhibition can be triggered automatically as well as result from voluntary suppression, it is always considered to be purposeful with regard to enhancing processing efficiency. *Retrieval-induced forgetting* denotes the phenomenon that selectively retrieving only a subset of information from memory lowers the accessibility of the non-retrieved rest of that set (Anderson, Bjork, & Bjork, 1994). This effect apparently reflects inhibition that occurs as a by-product of memory retrieval. Preceding research had shown that





retrieval can invoke interference by examining extensively such effects as the fan effect (Anderson, 1974), proactive interference (Underwood, 1948), retroactive interference (Müller & Pilzecker, 1900), or output interference (Smith, 1971). Retrieval-induced forgetting additionally suggests that an inhibitory mechanism serves to resolve interference arising during retrieval attempts. Typically, retrieval-induced forgetting is analyzed in a paradigm that consists of three main phases. In the learning phase, participants study several sets of items in combination with a shared (category-)cue that defines the specific set of items. In the subsequent retrieval-practice phase, participants are cued to recall half of the studied items from half of the sets. In the test phase, recall performance for all items is tested. Retrieval-induced forgetting manifests itself in significantly lower recall of non-practiced items from practiced sets as compared to non-practiced items from non-practiced sets. This effect has been demonstrated for a wide variety of materials, for example, different kinds of verbal materials (Anderson & Bell, 2001; Anderson et al., 1994; Carroll, Campbell-Ratcliffe, Murnane, & Perfect, 2007; Tempel & Frings, 2015a), images (Ciranni & Shimamura, 1999; Koutstaal, Schacter, Johnson, & Galluccio, 1999; Shaw, Bjork, & Handal, 1995), or motor actions (Tempel & Frings, 2013, 2014a, 2014b, 2015b; Tempel, Loran, & Frings, 2015). Several properties of retrieval-induced forgetting point to inhibition. First, only selective retrieval induces forgetting but retrieval-free kinds of selective practice (restudy) does not (Ciranni & Shimamura, 1999; Staudigl, Hanslmayr, & Bäuml, 2010; Tempel & Frings, 2016a). Second, the strength of interference predicts retrieval-induced forgetting (Chan, Erdman, & Davis, 2015; Tempel, Aslan, & Frings, 2016), whereas, third, its occurrence is independent from whether retrieval facilitates access to retrieved items (Hulbert, Shivde, & Anderson, 2012; Storm & Nestojko, 2010; Tempel & Frings, 2017). Fourth, retrieval-induced forgetting does not only emerge in tests using the same cues as during retrieval practice but also in tests probing





memory with independent cues (Anderson & Spellman, 1995; Weller, Anderson, Gómez-Ariza, & Bajo, 2013). Together, these properties indicate that the purpose of inhibition is the resolution of interference arising during retrieval attempts and that the observed forgetting effect is not the consequence of blocked access during the final test but reflects that individual item representations are affected. Blocking here refers to any theoretical explanation that assumes observed forgetting to result from previously strengthened traces getting in the way of retrieval routes. Blocking in this general sense involves changes in the relative amount of spreading activation arriving at traces of the forgotten items due to associative strengthening or weakening.

Inhibition can also result from voluntary interference control. The *think-no-think* paradigm examines effects of thought suppression (Anderson & Green, 2001). It also consists of three phases, starting with a learning and ending with a test phase. However, the second phase now requires participants either to recall or explicitly not to think about an item while they are presented with item-specific cues. The instructed thought suppression then entails an impaired accessibility to suppressed items in the test phase. This forgetting effect does not reflect blocking either. It regularly occurs even in the absence of any strengthening of previously to-be-recalled items. Thus, any impairment of previously suppressed items cannot be the result of practiced items blocking access to them. Moreover, the forgetting effect is cue independent, that is, it also emerges when a novel stimulus is provided as a cue in the test phase. For example, when participants are asked to recall a word from the learning phase that is an *insect* starting with the letter *r* after they had learned the word pair *ordeal – roach* and *ordeal* had been presented as a corresponding suppression cue for *roach* during the think-no-think phase. Again, this effect has been demonstrated for a variety of different materials, for example, words (Anderson & Green, 2001, Anderson et al., 2004; Hertel & Calcaterra, 2005), images (Depue, Curran, & Banich,





2007; De Vito & Fenske, 2017), and autobiographical memories (Küpper, Benoit, Dalgleish, & Anderson, 2014; Noreen & MacLeod, 2013; van Schie, Geraerts, & Anderson, 2013). The paradigm is an adaptation of the older go-no-go and the stop-signal tasks that are typically used to examine behavioral inhibition, that is, the voluntary stopping of a motor response (e.g. Logan & Cowan, 1984). It shows that we are able to recruit inhibitory mechanisms intentionally, controlling our ongoing cognitive processes.

　　*Directed forgetting* is another form of voluntary suppression. However, it does not pertain to precluding items from entering consciousness but it aims at deleting them from memory – at least, temporarily. In the list-method of directed forgetting, participants are asked to forget an initially learned item list before proceeding with encoding a second list. To their surprise, however, memory for both lists is assessed in a final test phase and compared with a control group that did not receive a forget instruction but was told to memorize and retain both lists. Typically, the forget instruction results in lower memory performance for the to-be-forgotten list 1 as compared to the control group but also in better memory for list 2. An inhibition of list 1 reduces interference between lists and, thus, facilitates access to list 2 (Bjork, 1989). Additionally, it has been demonstrated that directed forgetting can selectively affect only a part of previously encoded information (Delaney & Nghiem, 2009; Kliegl, Pastötter, & Bäuml, 2013). Therefore, mere mental segregation cannot account for costs and benefits of directed forgetting but this finding suggest targeted inhibition. In a related paradigm, the item method, several items are presented each followed by either an instruction to forget the just presented item or to retain it for a later test. Thus, the instructions to remember or forget items are interleaved within one item list (Basden & Basden, 1996). Typically, directed forgetting is beneficial here as well, enhancing memory for the to-be-retained information, that is, recall of those items is better than in a control





group not receiving any forget instruction. Both forms of directed forgetting have been demonstrated with a variety of materials, mostly words (for an overview see MacLeod, 1998) but also pictures (Hourihan, Ozubko, & MacLeod, 2009; Quinlan, Taylor, & Fawcett, 2010) or actions (Sahakyan & Foster, 2009; Tempel & Frings, 2016b).

## 2.2 Relation to activation

Inhibition increases processing efficiency because it lowers activation levels of cognitive representations that might get accessed via spreading activation because of strong associations to currently operating representations. However, access is precluded despite such strong associations when they are targeted by an inhibitory mechanism that serves to lower temporarily irrelevant information. Thus, inhibition sharpens the contrast between what now is important and what now is to be neglected. This might be achieved by impacting a single dimension of activation, that is, inhibition lowers access to memory representations because it subtracts a certain amount from the sum of spreading activation that the representation received through associations with stored knowledge and current percepts. Alternatively, there might be two coexisting levels of activation and inhibition that receive input via separate excitatory or, respectively, inhibitory links. The net accessibility then resulted from both the activation and inhibition system.

In any case, whether inhibition does occur depends on the strength of activation of competing information. If there is no competition, there is no need for inhibition. This general aspect applies to all inhibition effects. The think-no-think paradigm a priori involves an initial activation of items that makes them push to become aware but must be excluded actively from conscious recollection. Retrieval-induced forgetting only occurs if associations to a shared retrieval cue are strong enough for producing interference during retrieval attempts (Chan et al.,





2015; Tempel et al., 2016). Directed forgetting only weakens a to-be-forgotten list if a further item list is encoded, that is, if there is a potential for interference between lists (Pastötter & Bäuml, 2007), and stronger interference produces stronger directed-forgetting effects (Pastötter & Bäuml, 2010; Sahakyan & Goodmon, 2007). The dependence on activation of competing information, again, speaks of the adaptiveness of inhibition. Inhibition does not just by itself target some representations in memory and weakens them but it is recruited for weakening irrelevant information in order to facilitate accomplishing currently active task goals. Such goals may differ and comprise the encoding or recall of a new item set (directed forgetting), the targeted retrieval of related information (retrieval-induced forgetting), or explicitly precluding items from entering consciousness (think-no-think paradigm). It might seem that the latter example involves a kind of pure inhibition. However, it has been demonstrated that the strength of forgetting as a consequence of no-think-trials depends on the strategy to accomplish this demand, with distracting oneself by some means (e.g. thinking of some other items or recalling extra-experimental memories) entailing stronger forgetting (Hertel & Calcaterra, 2005). Thus, even inhibition during instructed suppression trials depends on activating alternative representations.

## 2.3 Links to contexts

Inhibition creates contexts. In cognitive psychology, the term context can either refer to external or mental contexts. External context comprises features of the present surroundings, for example the desk and working utensils in an office room, the blue sky outside the window, or traffic noise from a nearby street. Mental-context features may comprise, for example, thoughts of upcoming tasks, spontaneous memories, or a current mood state. Inhibition structures





cognition by mental contexts that may or may not be linked to the present surroundings.

Indeed, some theoretical models suggest that changes in mental context could be an alternative explanation of effects considered to reflect inhibition (Jonker, Seli, & MacLeod, 2015; Sahakyan & Kelley, 2002). However, the concepts are not mutually exclusive. In particular, inhibiting some information stored in memory segregates it from other representations, thus, establishing a separate set of items in memory or, if bound to already given set-defining features, it additionally deepens the contrast between sets. Context changes per se typically involve impaired accessibility in memory. A classic study by Godden and Baddeley (1975), for example, showed that divers that were tested on a word list they had learned under water were able to recall more words when tested under water as compared to being tested on land. The opposite was true as well. When tested on a word list they had learned on land, the divers were able to recall more words on land as compared to being tested under water. A variety of more subtle context manipulations have been examined since and also were found to affect memory performance with better recall if the test context matched the encoding context compared to non-matching conditions (Smith & Vela, 2001). Importantly, even mere mental context changes are able to impact memory (Sahakyan & Kelley, 2002). For example, the instruction to recall a specific event or scenery from autobiographic memory (e.g. imagining to walk through one's parents' home), impairs access to items that were encoded just before. Shifts in spontaneous thoughts or images popping into the stream of consciousness may occur naturally, also linked to external demands. Thus, a mental context change could be recruited to separate item sets and thereby foster organization in memory. In principle, such shifts do not necessitate assuming inhibitory mechanisms but they can also involve inhibition. In fact, most findings that have been interpreted as evidence of inhibition could be seen as resulting from mental context changes as





well. This does not speak against inhibition, however, because both might be linked with each other, especially, if inhibiting a set of items is regarded as an act of segregation in memory always involving a mental context change. Demonstrations of the cue independence of retrieval-induced forgetting and forgetting effects in the think-no-think paradigm suggest that mental context changes are in fact a consequence of inhibition but that inhibition persist in novel contexts because independent probes minimize the overlap of task features between encoding and recall trials. Moreover, inhibition effects also have been found with implicit memory tests that preclude recall attempts in which context representations could be used as retrieval routes (e.g. Perfect, Moulin, Conway, & Perry, 2002; Veling & Van Knippenberg, 2004). Taken together, assuming mental context changes is not able to account for the full range of inhibition effects that have been reported. Rather, mental context changes seem to occur in concert with inhibitory processes.

## 2.4 Temporariness

Inhibition is not permanent. In fact, its temporary nature is a crucial aspect of its adaptiveness. Inhibition weakens information that is currently irrelevant but it does not completely erase it from memory. When information becomes relevant again for a new task or in a new context, persisting inhibition would be maladaptive. Apparently, it is not however. Inhibition vanishes after it has served its purpose. Suitable cues are able to trigger a release from inhibition that even might occur as a mere product of passing time. Correspondingly, inhibition effects in memory have been found to vanish after a day or a week (MacLeod & Macrae, 2001; Nørby, S., Lange, M., & Larsen, A. 2010; Storm, Bjork, & Bjork, 2012) and when features of the encoding context are reinstated that trigger a release from inhibition (Jonker et al., 2015; Sahakyan & Kelley, 2002). However, relative to the total number of studies demonstrating





inhibition effects, investigations on their duration or context-induced releases from inhibition are rather rare. Although temporariness is well documented, it is hardly possible to predict the specific persistence, therefore.

Depending on the precise representation that is targeted by inhibition, relevant context features may differ for triggering a release from inhibition. In retrieval-induced forgetting and the think-no-think paradigm, inhibition affects representations of individual items in order to preclude them from entering consciousness when appropriate retrieval cues are present. In contrast, list-method directed forgetting involves inhibition of the representation of a whole set of items instead of single items. Thus, providing copy cues in a recognition test (i.e. the very stimuli that have been encoded before) still results in forgetting effects when individual items had been inhibited (e.g. Hicks & Starns, 2004; Racsmány, Conway, Garab, & Nagymáté, 2008; Spitzer & Bäuml, 2007; Veling & Van Knippenberg, 2004) but triggers a release from inhibition when the representation of an item set had been inhibited because access occurs independently from this inhibited representation (Geiselman, Bjork, & Fishman, 1983).

### 3. Inhibition in Computer Science: Learning from cognition

In many areas, the way human mind and body work has shaped the development of new technologies, either to support humans or to take inspiration from them (e.g., Human Computer Interaction, Booth, 2014; Augmented Reality, Billinghurst, Vlark, & Lee, 2015; Neural Networks, Graupe, 2013). Here, we particularly focus on incorporating the principle of cognitive inhibition in Personal Information Management (PIM). PIM refers to the practice and study of activities performed by people to acquire, organize, maintain and retrieve information for everyday use (Jones, 2007). In PIM, first discussions on psychological issues in the field date back to Lansdale (1988; for more recent considerations see Elsweiler et al., 2008; Vavoula &





Sharples, 2009). The most recurrent insight borrowed from Cognitive Psychology is the episodic and associative nature of human memory as well as the important role of context: when trying to recall or retrieve previously stored information (documents), contextual information captured at storage time can facilitate the retrieval process. Based on this assumption, many systems for personal information management and search (Lamming & Newman, 1992; Lansdale & Edmonds, 1992; Dumais et al., 2016; Chen et al., 2009) treat contextual information of a given file (such as creation time and location, involved people, other accessed files, co-occurrent events) as memory cues and exploit them for search.

However, we see context in a much broader sense, especially inspired by the notion of context from Cognition. Considering approaches using context in Computer Science, one finds different views on what context is comprised of as well as context models that are influenced by the respective area of research (see, e.g., overviews in Brézillon, 1999; Dourish, 2004; Hoseini-Tabatabaei et al., 2013; CONTEXT conference series, e.g. CONTEXT 2017, Brézillon, Turner, & Penco, 2017). Those approaches taking context serious provide a clear definition on how context is defined as well as clarify its role in their systems by providing formal models representing context to acquire, store, identify, compare, retrieve and reason about contexts.

In general, context in Computer Science is often used to deal with the situation surrounding and influencing a goal, task or object in focus which being a priori not available and usually acquired during runtime from the physical environment or available information space. Such context is used to characterize the situation and to contribute to the understanding of the goal/task/object in focus as defined by Dey (2001): "Context is any information that can be used to characterize the situation of an entity. An entity is a person, place, or object that is considered relevant to the interaction between a user and an application, including the user and applications





themselves." (p. 4)

Treating context as an associative set of concepts or information items and being used to allow a better interpretation of a situation and act accordingly, i.e., context-aware, is another successful adoption of insights from Cognition. For instance, the famous CYC approach with its rich dimensions to express context as a region in an n-dimensional space to represent common knowledge in formal reasoning (Lenat, 1998). Close to our view on context is the approach taken by Schwarz (2010). There, contexts were each represented by a probability distribution of concepts and their probability being in and influencing a specific context. The concepts originate from a personalized semantic network maintained by a service running on the user's computer. This service also observes the user's desktop activities and treats these observations as evidences to adapt the contextual probability distributions using Bayesian Inferences. This allows for context-aware applications such as a computer desktop allowing to switch between contexts. The interface is then adapted accordingly by providing the application windows which were active in that context, thus also hiding non-relevant ones (Schwarz, Kiesel, & van Elst, 2008).

The associative nature of human memory also inspired the works by Katifori et al. (2010) and Tran et al. (2016), which identify task-related, currently relevant information items by tracking usage activity and propagating such signals of importance or activation over semantically related resources. In particular, Tran et al. (2016) model a short-term forgetting process by applying a negative exponential decay to the importance of accessed resources, which nevertheless can become important again when semantically related resources are accessed. This work was conducted as part of a bigger project aiming at establishing a short- and long-term information management inspired by the selective forgetting mechanisms in the human memory (Niederée et al., 2015). Finally, instead of dealing with on-demand recall and searches, Rhodes &





Starner (1996) presented an autonomous agent to augment human memory by continuously searching for potentially useful information based on the current user activity. Such an agent handles situations where users do not remember enough to realize that they have forgotten something in the first place.

Whereas such software typically uses a semantic-network structure and adopts the principle of spreading activation, an explicit consideration of inhibitory mechanisms often is neglected although some principles of such software in fact do resemble inhibitory processes. In the following, we will give an overview of existing software already incorporating inhibitory mechanisms and additionally outline paths for future development. We investigate the concept of inhibition in Computer Science along the same four functional aspects discussed in human cognition.

### 3.1 Processing efficiency

Dealing with large quantities of information and information overload are crucial challenges in Computer Science and, in the age of digitization and the internet, are gaining even more in importance (Bawden & Robinson, 2009). Thus, there is active research in information-technology methods that help users in better dealing with the large amounts of information such as methods for supporting effective search.

Looking into such methods, it can be observed that there is a preference for propagating useful information, while inhibiting unwanted information is not in the focus of most methods for dealing with information overload. Information Retrieval (Manning et al., 2008), for example, which is the primary access method for large information spaces such as the Web, relies on the idea of finding the most important or relevant information in an information space given an information need (typically articulated as a search query) and ranking them in order of relevance.





Also Recommender Systems (Goldberg et al., 1992), which pro-actively suggest information to the user and are very popular in the commercial context, follow the principle of selecting the most promising information (e.g. based on what "similar" users liked) and providing it to the user. This also explains the popularity of the spreading activation concept as an adoption of cognitive science concepts in Computer Science (Crestani & Lee, 2000; Rocha et al., 2004; Katifori et al., 2010), where relationships between information items are exploited for better estimating their importance and the dynamic changes in importance.

When methods in Computer Science additionally made explicit use of inhibitory mechanisms, the motivation for introducing forms of inhibition always has been related to efficiency and/or protection, thus, following the idea of adaptive inhibition in cognitive theories. Efficiency here refers to improving machine efficiency as well as human efficiency, e.g., by avoiding to loose focus or putting attention to unimportant things. When performing tasks with a computer, interruptions triggered by messengers, emails, etc. are an important issue for human attention; their negative impact on task performance has, for example, been thoroughly investigated (e.g. Adamczyk & Bailey, 2004; Horvitz et al., 2001; Horvitz & Apacible, 2003). Inhibitory methods such as temporarily blocking selected applications or all types of system messages are today a common practice to reduce interruptions. They are supported by many applications such as Skype and also by operating systems. Another inhibition-based method supporting human efficiency, is spam filtering, where especially filtering junk or spam in email messages (see Blanzieri & Bryl, 2008 for an overview) is important for personal information management. In email spam filtering useless content is identified and suppressed, e.g., not shown in the incoming mails, sorted out or marked for not wasting attention to it. Typically machine learning is used for classifying email messages into legitimate messages and spam (Sahami et al.,





1998; Sculley & Wachman, 2007). On a more generic level, there are also methods of information visualization, which apply inhibition by hiding part of the information, e.g., by applying hierarchical aggregation (Elmqvist & Fekete, 2010), thus hiding unnecessary detail. The importance of adequate and selective information visualization has already been recognized in the 1990s (Shneiderman, 1996) and is even more important now in the age of Big Data.

In addition to improving human efficiency, inhibition-based methods are also used for increasing computing efficiency. One example for such methods is index pruning (Carmel et al., 2001), where entries, which only have a minor effect on retrieval results, are deleted from the information retrieval index. The retrieval index is an auxiliary data structure, which ensures the efficiency of information retrieval methods. By index pruning, index size is considerably decreased easing its storage and processing. Another inhibitory method for improving computing efficiency is memory caching. In such fast but limited memory structures less popular content is swamped out in favor of more important or more frequently used content. Memory caching is core for the efficiency of technologies such as Web search engines (Fagni et al., 2006; Baeza-Yates et al., 2007) and can also be combined with index pruning (Skobeltsyn et al., 2008).

Protection of information is a second important purpose for introducing inhibitory methods in Computer Science. This, for example, includes the socially-driven forms of inhibition in Social Media platforms such as Facebook, where access to information is restricted e.g. based on group membership (see e.g. Liu et al., 2011) as well as many other forms of access control.

### 3.2 Relation to activation

*Activation* and *inhibition* are dual concepts. The former means making (temporary) important information items vivid and accessible, while the latter is about ignoring the unimportant ones. We will discuss different ways of combining activation and inhibition. Before





delving into their description, we first introduce how short-term importance and activation

mechanisms can be modeled in systems for personal information management.

**Table 1: Common time-decay functions used in literature (Tran et al., 2016).**

| Name | Function | Parameters |
|------|----------|------------|
| Polynomial Decay | $\dfrac{1}{(t - t_a)^\alpha + 1}$ | $\alpha$: decay rate |
| Ebbinghaus Curve | $e^{\frac{(t_a - t)}{s}}$ | $s$ : relative memory strength |
| Weibull Distribution | $e^{\frac{-\alpha\,(t - t_a)^s}{s}}$ | $s$ : forgetting steepness $\alpha$ : remembering volume |

The importance of an information item (e.g. textual document, image, presentation, etc.)

with respect to the user attention is often modeled based on its usage activity. The activation level

is decreased over time to simulate forgetting behaviors (Katifori et al., 2010; Tran et al., 2016). In

particular, Tran et al. introduce the term *memory buoyancy* to represent how much an item is

vivid in the short-term memory of the user (still estimated via usage activity). We adopt this

notion for the rest of our discussion. The basic building blocks are time-decay functions like

those listed in Table 1, where $t$ is the current time point and $t_a$ is the last time the given resource

was accessed. The parameters of such functions regulate how fast the decrease of importance is,

with numerical values that should be identified based on the characteristics of the particular

scenario at hand.

Spreading activation mechanisms are then applied to raise the memory buoyancy of a





resource when other related ones are accessed. An example is given in Figure 1 (left), which

shows different temporal behaviors of the memory buoyancy of a given resource. The continuous

line represents the spontaneous decay in case the resource was only accessed once at the

beginning of the curve. However, the access and usage of semantically related resources (e.g.

belonging to the same event or involving the same persons) would activate the "remembering" of

the resource under consideration, making its memory buoyancy higher. This is the case

represented by the dashed line of Figure 1 (left), where the peaks represent time points when

related resources were accessed.

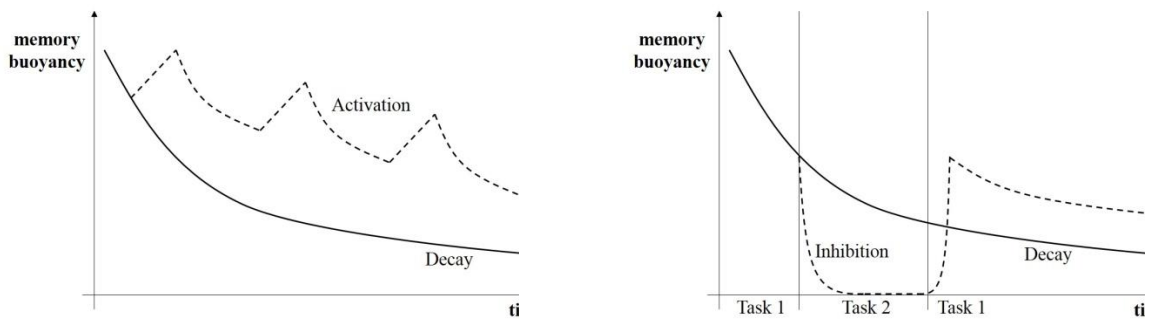

**Figure 1: activation (left) and inhibition (right) within pure decay-based model of task-related relevance.**

Figure 2 presents an implementation of memory buoyancy based on the works of (Tran et

al., 2016; Jilek et al., 2016). The memory buoyancy curves show activation and decay of

concepts of a user's personal semantic network over a period of time. Evidences are derived from

direct user interaction such as accessing a concept (here, working on the tasks "Extended

Abstract" and "Chapter"), influenced via spreading activation over semantically related resources

(here, the project ForgetIT, being a topic of the tasks), or by semantic triggers such as completing





a task (finishing the task "Extended abstract" leads to an explicit forgetting by decreasing its memory buoyancy).

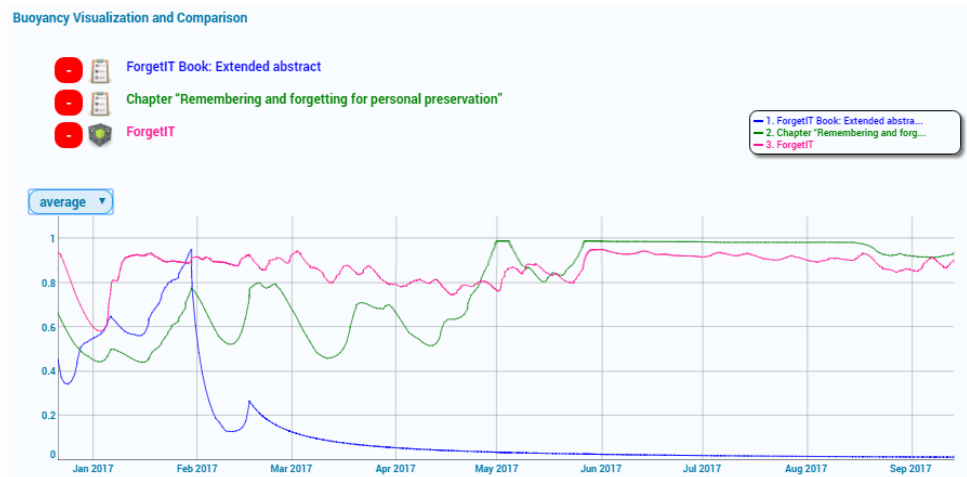

**Figure 2: Memory buoyancy graphs of concepts in the Semantic Desktop system presented in (Jilek et al. 2016).**

*Information Hiding.* An aforementioned inhibition-based method consists in hiding those information items that are currently not important. Being irrelevant and interfering information hidden from the view of the user (i.e. inhibited), he or she could stay more efficient and focused towards the goal with a decluttered information space. An important scenario for combining activation and inhibition is created, when combining information hiding with search. Search as an activation-oriented mechanism, which highlights information relevant to a user's information need (see above), can be combined with the inhibition-oriented mechanism of information hiding exploiting the concept of memory buoyancy described above: only those relevant resources (identified by the search method) with "sufficiently" high values of memory buoyancy would be shown in a search result. This can be achieved through a *hiding* function that uses a threshold for





deciding what to hide:

$$hiding(r,t) = \begin{cases} 1 & if \quad mb(r,t) < \tau \\ 0 & otherwise \end{cases} , \ 0 \leq \tau \leq 1$$

where *mb(r,t)* is the memory buoyancy of resource *r* at time *t* and $\tau$ is a given threshold that regulates the amount of resources to be hidden. Thus hiding would act as an information filter here. Clearly there is not a unique value of $\tau$: it should be tuned based, for instance, on the actual application domain and user preferences.

When listing search results, we use binary hiding decisions. In other scenarios (e.g. on the desktop), it is also possible to smoothly hide resources (their icons) by varying their transparency according to their memory buoyancy, for instance:

$$hiding(r,t) = 1 - mb(r,t)$$

where $0 \leq hiding(r,t) \leq 1$ is the degree of transparency of a resource. According to this, the less a resource is accessed the more it will become transparent. In the above equation we have used a linear dependency between memory buoyancy and hiding (transparency), but other types of relations could be investigated as well.

Figure 3 shows an example from the Semantic Desktop system: different views on the same event and its attached information items are shown over a period of time. The first image shows the event while it was taking place with its multitude of items such as attendees, locations, e-mails, tasks and presentations.  The next two images show how items are hidden over time. This is because their memory buoyancy value drops below a certain threshold (here 0.5 in a range from 0 to 1). If users are still interested in that part they have to explicitly request the forgotten





items by pressing the button 'show forgotten' which then lists all connected items again.

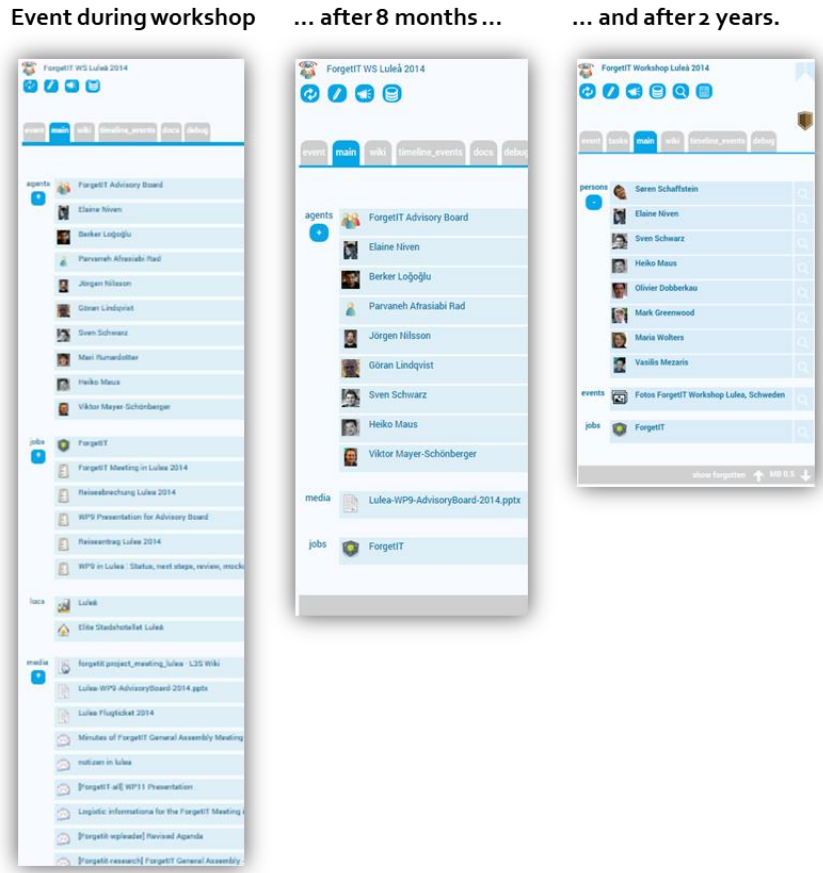

**Figure 3: Using memory buoyancy for information hiding in the Semantic Desktop system: information items with low memory buoyancy value are hidden from direct view of a user when browsing the semantic network.**

*Ranking.* A more advanced way of combining search and inhibition is achieved, when incorporating inhibition mechanism with ranking. As mentioned before, ranking consists in sorting a set of documents that match a user query according to how much they are relevant to the query itself (which encapsulates an information need of the user). Ranking is important because





users usually "digest" only a limited number of results, typically the first ones shown, due to drop of attention or limited time. After ranking, the highly relevant documents (according to the search method) are on top of the list and, therefore, more visible and prominent to the user.

While this indirectly implies that irrelevant documents get inhibited by staying at the bottom of the search result list, we want to integrate inhibitory contributions into ranking functions more explicitly. We propose to realize this by exploiting usage activity, its semantic propagation, and any other contextual information that might be available as it is incorporated in memory buoyancy. When searching, we would like to penalize documents with low memory-buoyancy values in the ranking, similarly to how was discussed for the case of information hiding. The way of combining search information hiding and search can be considered as a very simple form of doing this.

However, there might be other situations where the pure relevance of a resource should be considered as well. For instance, if a document is really relevant to the query), then the user might still need/want it irrespectively of when it was lastly used. This case, especially, encompasses all those situations when we are looking for one specific resource, which we cannot find indeed because we forgot its location due to not using it.

More generally, we would like the ranking to incorporate both relevance and inhibition and to properly balance them. Assuming the *relevance* and *inhibition* functions can be computed at time $t$ given a query $q$ and for each retrieved resource $r$, then a simple linear ranking function combining relevance and inhibition is the following:

$$score(r, q, t) = f\big(relevance(r,q), inhibition(r,t)\big) = \alpha \cdot relevance(r,q) - \beta \cdot inhibition(r,t)$$

$$\alpha + \beta = 1$$

where relevance pushes items up, while inhibition turns them down. The $\alpha, \beta$ parameters regulate





the contributions of these two aspects and should be identified based on the characteristics of the application scenario and user preferences. Query-document relevance can be computed, for instance, based on standard measures such as *tf - idf*, *bm25*, *language modeling* (Manning et al., 2008). Inhibition can be modeled based on the memory-buoyancy value, but other inhibitory behaviors referring to specific scenarios and exploiting particular information at hand could be plugged in as well. An inhibitory contribution would stay applicable also when moving away from personal information search, for instance in web search: Inhibition can be modeled in this case based on search history, collective view counts of web pages over time, as well as popularity trends of topics related to the queries.

### 3.3. Links to contexts

Defining task context in Personal Information Management might be one of the prime ways to incorporate inhibition in computer systems. Tran et al. (2016) followed this approach by linking task switches to sudden drops in the memory buoyancy of information units that had been relevant to a preceding task but were unrelated to the current task (Figure 1, right). At some point in time, the user stops working on Task 1 and starts Task 2, which is semantically very different from the previous one. The solid line illustrates the slow decrease of memory buoyancy due to non-use and non-activation. Incorporating inhibition, the importance level is temporarily decreased at a much faster rate (dashed line) so that the resource quickly gets discarded for the new task. In case the user goes back to Task 1, the inhibitory effect would vanish and the memory buoyancy would be reset to its former value. This also shows the *temporariness* of inhibition (see below). Note that the patterns shown in Figure 1 are meant for exemplification purposes only: the actual amount of inhibition depends on different aspects such as, for instance, the characteristics of the handled resources as well as the temporal and semantic gap between the





tasks involved in the switch. This inhibitory mechanism is very similar to the kind of inhibition presumably underlying directed forgetting in the human mind. In both, context segregates two sets of information and inhibition precludes the irrelevant set from interfering while the relevant set is operated with.

In Personal Information Management two types of inhibitory mechanisms can serve this purpose: i) inhibitory mechanisms for quickly abandoning the resources only relevant to the previous task and not to the new one (context switching support) and ii) inhibitory mechanisms that prevent activation of resources outside the current context (context limiting support). The granularity of "context" may range from rather implicit short-term contexts like search queries to explicit long-term contexts associated with tasks. Within this spectrum are also medium-term contexts or sub-contexts etc. Information hiding and ranking as described above can be enhanced by making use of information about contexts for deciding which information to show to the user. Furthermore, resources (weakly or strongly) associated with a task could be ranked by taking into account additional information like closeness to inhibited contexts, etc. As mentioned above, a prerequisite for the inhibitory methods is that we know which information items are related to a specific context. This can be achieved by users giving explicit feedback as well as by background applications tracking and analyzing user activities (Biedert, Schwarz & Roth-Berghofer, 2008; Maus et al., 2011). If activation crosses the borders of the current context, inhibition comes into play to prevent activation of items outside the current context. This approach follows the basic idea of inhibition: For a certain cue several task contexts could be relevant – the more similar they are the higher is the possibility of interference. Such interference is prevented by applying inhibitory measures causing some items to be suppressed while a certain task context is active.

*Implementation in Semantic Desktop.* The first type of inhibitory mechanism has already





been implemented in the context of the Semantic Desktop in (Schwarz, 2010; Maus et al. 2011).

In the following, we discuss ideas for realizing the second mechanism, Context Limiting Support,

in more detail. Basically, we enhance Spreading Activation algorithms (that have also been used

for implementing the previously mentioned memory buoyancy, see Figure 2) with inhibitory

effects. As a prerequisite, we assume that users are aware of the concept "context" (Gomez-Perez

et al., 2009) and that they successfully/willingly organize their data accordingly, e.g. information

items are stored using a personal information model (PIMO) (Sauermann et al., 2007; Maus et al.,

2013), which reflects a user's mental model in a semantic network and is based on the so-called

Semantic Desktop approach (Sauermann et al., 2005). Thus, all things belonging to the same

context in the user's mind – calendar events, files, emails, bookmarks, topics etc. (see Figure 3

for an example) – are actually stored within a corresponding context in their PIMO. For example,

a context "Vacation in Spain 2017" could contain the calendar entries belonging to this holiday

trip as well as files (tickets, photos, …), emails (booking, holiday greetings, …) and bookmarks

(offers from various travelling sites, interesting places to visit, …).

　　　Switching to another task means switching to another context and, hence, causes the

corresponding areas of the semantic network in the background (i.e. the user's PIMO) to be

stimulated. Figure 4 illustrates this: It shows a semantic graph consisting of information items –

also called things – (depicted as nodes) and relations between them (edges). Additionally, three

areas corresponding to three different contexts (of tasks in this example) are highlighted with a

blue, yellow and red circle. Technically, contexts are subsets of (connected) semantic nodes in

the graph. Contexts are either defined explicitly by specifying the enclosed nodes, or they emerge





as graph extensions originating from specified nucleus nodes.

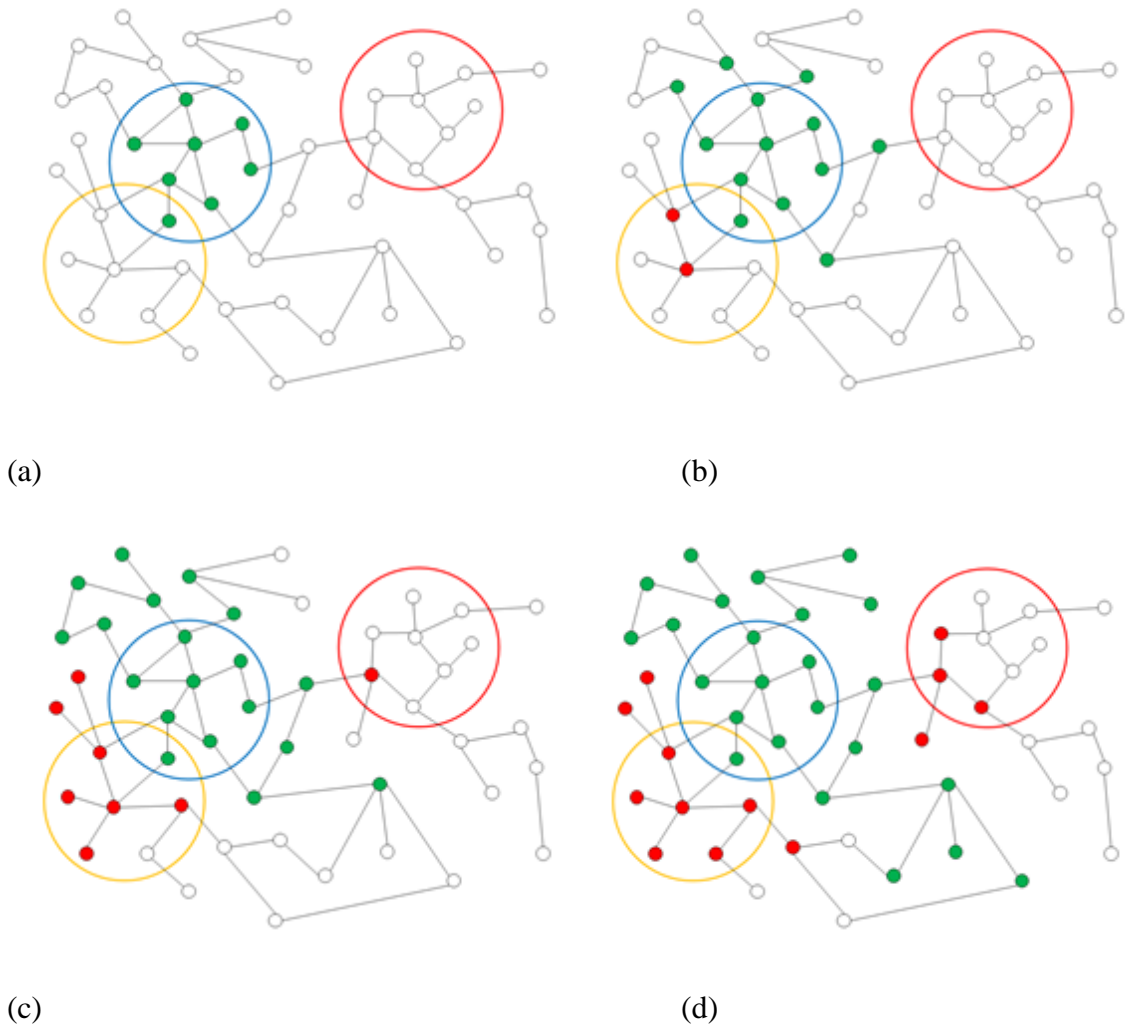

(a)                                                     (b)

(c)                                                     (d)

**Figure 4: Spreading Activation with inhibitory effects in a semantic graph.**

In the first image (a) things explicitly belonging to the selected task are activated (green nodes in the blue circle). We see that there is one activated thing in the overlapping area of the blue and the yellow circle, meaning it is part of both task contexts, e.g. a shared topic, person, or document. In the second image (b) the activation starts to spread across the network also





stimulating those things that are not directly confirmed to be in the blue context but are strongly associated with it (green nodes outside the circle). Traditional spreading activation would also activate the red nodes. But since they are explicitly belonging to a different context (associated with the yellow task), we do not want them to interfere with the currently selected context (the blue one). Therefore, they are inhibited (i.e. not activated here). In (c) and (d) the activation spreads even further across the network, again leaving out those things either directly belonging to another task or if they can only be reached by crossing another context.

## 3.4 Temporariness

Inhibition is not permanent in human memory. Therefore, neither should inhibitive computational methods be. Inhibition serves to decrease the activation under certain circumstances for a limited period of time. Pure time lapsing is the most obvious choice when referring to temporal aspects, as it is responsible of document aging (losing relevance over time) and potential loss of sharpness in human memory, However, it makes sense to complement pure aging with usage-related information. Older, but recently accessed documents should stay vivid in the digital working memory, while unused ones (even if more recent) could be inhibited. Implementing inhibition linked to task context in Personal Information Management matches exactly this requirement by only keeping up inhibition of task-unrelated information items as long as work on the current task continues.

More generally, similar approaches have been adopted in the exploitation of usage data and behaviors in a wide range of problems, such as, for predicting short-term user interest (White et al., 2010) and surfing behaviors (Awad et al., 2008) during web searches, for caching and prefetching query results (Fagni et al., 2006), for query log mining (Silvestri, 2009), for predicting upcoming user actions in different domains (Fitchett & Cockburn, 2012), for





recommending web pages (Nguyen et al., 2014) or personal documents (Rhodes & Starner, 1996). Regarding personal information management, resource usage activity has been leveraged for deriving retrieval cues to ease personal document re-finding (Lamming & Newman, 1992; Dumais et al., 2016) as well as for mining temporal semantic relationships between documents (Soules & Ganger, 2005; Chen et al., 2009; Katifori et al., 2010; Tran et al., 2016). This last group of works is of particular interest for the present article, because (i) they estimate the temporary relevance of resources to a current task based on their usage activity over time and (ii) they propagate such relevance by "activating" semantically related information items, although they might not have been accessed explicitly. While these approaches traditionally focus on resource importance/relevance instead, recent developments open them to an integration of inhibition, as described above. Their explicit notions of temporariness and semantic associations/propagations enable a broad application of inhibitory mechanisms in these systems.

## 4. Concluding remarks

We reviewed studies documenting a variety of effects speaking of inhibition in human memory and described recent efforts of incorporating the principle of cognitive inhibition in computer systems, particularly regarding Personal Information Management. For future developments, we believe that a particularly promising approach will be to continue linking contexts in Personal Information Management with inhibitory mechanisms. Figure 5 depicts a general approach that may guide such efforts of combining insights from Cognitive Psychology with applications in computer systems and their effects on computing efficiency as well as supporting the user's mental processes. A current focus of the work on contexts is on automatization of context creation, that is, requiring fewer explicit categorizations by users but identifying which context a user is working in by his or her actions. This implies that the





computer recognizes when active tasks switch and supports such switches with changing the context within, for example, the Semantic Desktop. Also, novel contexts might be suggested to the user when his or her actions do not match an already existing context because a new task has been begun. Inhibition can facilitate recognizing context switches as well as the creation of novel contexts because it intensifies the contrast between active an inactive contexts. The targeted access of documents from a currently inhibited context indicates a potential task switch more reliably because there is a sharper contrast between an active context and an inactive and inhibited context than between active and inactive contexts not affected by inhibition. Thus, task switches are detected more easily. Novel contexts could be suggested when a number of documents (or other elements) from different currently inhibited contexts are accessed, hence, the access pattern does not match any stored context. If contexts are only defined for things that had been relevant for a certain task before but are inhibited because they are irrelevant during work on a current task, access of inhibited items from different contexts signals that they become relevant again but for a novel task. In contrast, if previously relevant items were not inhibited, they would be treated in the same manner as anything that had been relevant at some point before, irrespective of the context, and their activation differed to the same amount from activation of things that never had been relevant for any task so far. Crucially, the concept of inhibition implies that processing efficiency is increased by temporarily weakening irrelevant information that might interfere with current intentions precisely because it had been relevant before and may become relevant again. Therefore, accessing an inhibited item is important information for a context switch and accessing it in the absence of a pattern corresponding to a





stored context suggests the creation of a novel context.

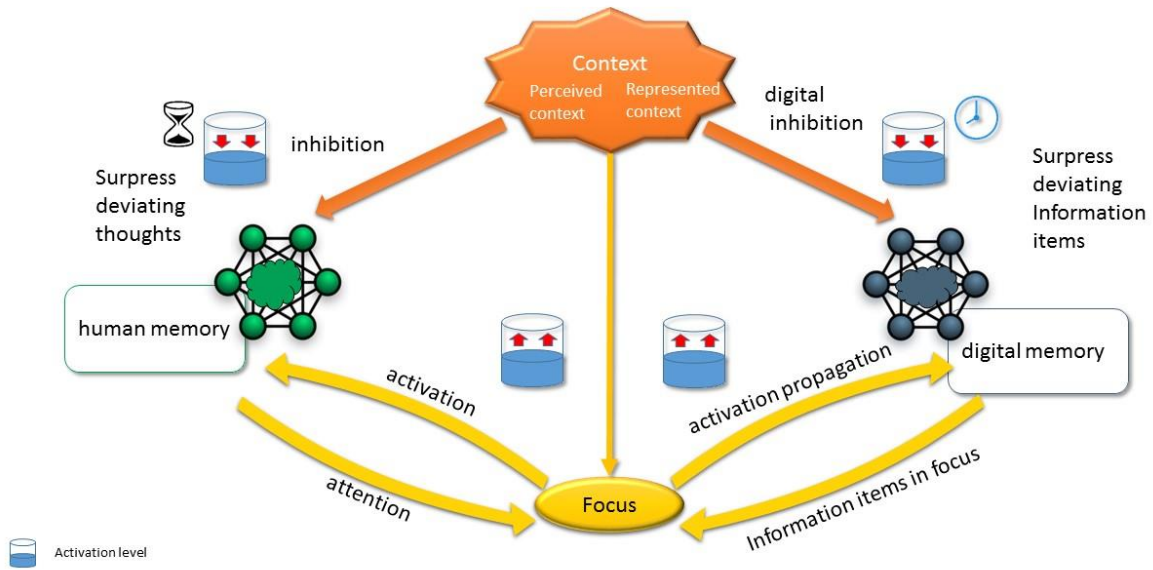

**Figure 5: Model of a combined psychological-computer-science approach for investigating inhibition in human and digital memory.**

In the past, Cognitive Psychology mainly has been inspired by Computer Science, not vice versa. Cognitive Psychology partly owes its existence as a scientific discipline to Computer Science. The computer served as an important metaphor during the cognitive revolution. The transition from behaviorism to cognitive models as the leading theoretical framework in scientific psychology was enabled by adopting information-processing ideas from Computer Science. Distinctions, for example, between hard- and software or between storage units and working units have been highly valuable stimulations for cognitive theorizing. Inhibition, however, is no processing principle of computers, whereas, as explained above, psychological research of the last decades has documented much evidence for inhibition to be a processing quality on its own





in cognition. We believe that Computer Science today benefits from incorporating this quality into their models. The implementations described in the present article are only a beginning, yet promising. They strongly suggest that inhibitory mechanisms can be used to enhance efficiency in human users as well as the efficiency of machine processing. Cognitive Psychology finally can give something back. We have learned much about the human mind by applying the computer metaphor. Now, it is time for computers to learn from the human mind.